\documentclass[11pt]{article}

\usepackage{graphicx}
\usepackage{url}
\usepackage{hyperref}
\usepackage[numbers,sort&compress]{natbib}
\usepackage{fancyhdr}
\usepackage{colm2024_conference}
\let\ours\relax
\usepackage[utf8]{inputenc}
\usepackage[T1]{fontenc}
\usepackage{booktabs}
\usepackage{amssymb}
\usepackage{amsmath}
\usepackage{colortbl}
\usepackage{tabularx}
\usepackage{array}
\usepackage{geometry}
\usepackage{xspace}
\newcommand{\ours}{XXX-Image\xspace}
\usepackage{enumitem}
\usepackage{float}
\usepackage{multirow}
\usepackage{makecell}
\usepackage[table]{xcolor}
\usepackage{threeparttable}

\newcolumntype{L}{>{\raggedright\arraybackslash}X}
\newcolumntype{C}{>{\centering\arraybackslash}X}
\newcolumntype{P}[1]{>{\raggedright\arraybackslash}p{#1}}
\newcolumntype{S}[1]{>{\centering\arraybackslash}m{#1}}

\definecolor{darkblue}{rgb}{0,0,0.5}
\definecolor{skyblue}{rgb}{0.289,0.484,0.72}
\definecolor{myred}{RGB}{190,0,0}
\definecolor{lightgray}{RGB}{245,245,245}
\hypersetup{
  colorlinks=true,
  citecolor=skyblue,
  linkcolor=red,
  urlcolor=darkblue,
  hypertexnames=false,
  bookmarksopen=false,
  bookmarksnumbered=true
}

\title{
From Corpora to Co-Evolving Capabilities:\\Capability-Centric Data Design for Generalist Image Generation}
\author{%
Xingjian Wang$^{*}$, Zhao Wang$^{*}$, Taihang Hu$^{*}$, Jun Zheng$^{*}$, Zhengrui Chen$^{*}$, Qinye Zhou$^{*}$, Zhengtao Wu$^{*}$, Yongchao Du, Zuan Gao, Chao Lin, Yefeng Shen, Yuan Wang, Xiaoli Xu, Zhengze Xu, Hao Yan, Denghui Yang, Yuhang Yu, Huayu Zhang, Mingzhou Zhang, Mengting Chen$^{\dagger}$\\
{\small $^{*}$Equal contribution. \quad $^{\dagger}$Corresponding author.}\\
{\textbf{Alibaba Group}}\\
}
\date{\today}

\makeatletter
\def\@maketitle{%
  \vbox{\hsize\textwidth
    {\centering{\Large\bfseries\@title\par}}
    \vspace{-0.3cm}
    \begin{quote}\rule{\z@}{6pt}{\centering\par\@author\par}\end{quote}
    \vskip 0.3in minus 0.1in
  }%
  \thispagestyle{plain}%
}
\makeatother

\begin{document}
\pagestyle{plain}
\maketitle

\begin{abstract}
Large-scale image generation has benefited from advances in data scale, quality, rebalancing, and recaptioning, yet conventional pipelines typically optimize task-specific datasets in isolation.
A central challenge is not only how to curate each task-specific corpus, but also how to organize heterogeneous supervision according to the dependencies among generative capabilities.
We present a \textbf{capability-driven data infrastructure} that couples capability-specific supervision construction with capability-aligned curriculum scheduling. 
Its three specialized yet interoperable data engines build complementary relational supervision for text-image grounding, inter-image transformation, and image-knowledge association, while caption experts align T2I and editing supervision across tasks and granularities. 
A multi-stage curriculum jointly evolves task composition, visual-concept distribution, data quality, and image resolution along the dependency order of capability acquisition, with capability-aware evaluation closing the loop through targeted retrieval, expert construction, and gap-aware resampling. 
At scale, the framework curates a 440M-image T2I corpus, 120M editing pairs, and over 27M image-entity pairs. 
With this infrastructure, we train multimodal diffusion models at two scales from scratch, with 3B and 6B sizes respectively. 
We conduct quantitative evaluation on CPI-Bench, along with qualitative evaluations across diverse text-to-image and editing scenarios.
Experimental results present broad visual coverage, versatile rendering, and effective transfer across generative capabilities.
\end{abstract}

\section{Introduction}
\label{sec:introduction}
Recent advances in image generation, spanning both text-to-image (T2I) synthesis and image-to-image (I2I) transformation, have been accompanied by a parallel evolution in data construction.
Recent progress in diffusion models has been supported by increasingly large datasets such as LAION-5B, COYO-700M, and MMC4~\citep{laion,coyo,mmc4}.
Furthermore, some studies have shown that data quality, semantic diversity, and caption density materially affect alignment and learning efficiency~\citep{li2024scalingt2i,chen2026scalingtext}.
Therefore, modern data pipelines like Qwen-Image~\citep{wu2025qwen} and Seedream~\citep{seedream2025seedream} invest heavily in expanding data coverage, filtering, semantic balancing, and recaptioning.
These advances answer how to construct a better corpus in aggregate, but leave a different question underexplored, i.e., \textbf{\emph{how should data be organized to develop a collection of interdependent generative capabilities~?}}

This question becomes central for generalist image generators.
Different generative capabilities do not emerge simultaneously from scratch~\citep{bagel2025}.
Instead, they develop in a clear dependency order, which closely aligns with the stages of the model's curriculum learning.
For example, semantic alignment of T2I data provides reusable concepts for structured generation and image editing~\citep{xia2025dreamomni}, and coarse or simple content provides a foundation for learning at higher resolutions and with more complex structures~\citep{cai2025z,team2025longcat,team2026firered}.
Consequently, the utility of a training sample depends not only on its quality, but also on the capability it targets and its intrinsic relationships with other samples.
Many conventional data pipelines treat datasets as task-specific design units, such as image-caption pairs for T2I or source-target pairs for editing, and optimize them largely in isolation~\citep{dalle3,hui2024hq,pico-banana-400k}, which can limit supervision sharing across tasks.

\begin{figure}[t]
\centering
  \includegraphics[width=1.0\textwidth]{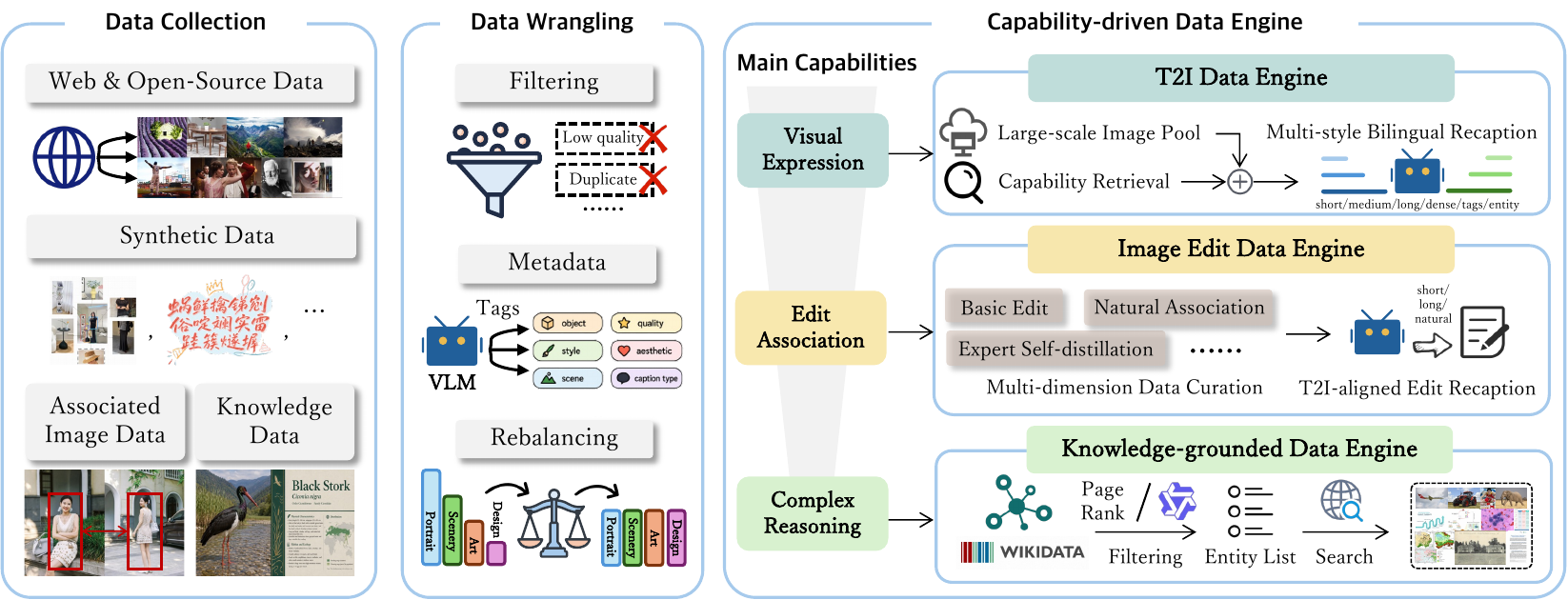}
  \caption{Capability-specific data construction in our framework. Shared collection and wrangling feed three specialized but interoperable engines for visual expression, editing association, and knowledge-grounded reasoning.}
  \label{fig:data_framework}
\end{figure}

To address this issue, we treat data curation as a capability-driven infrastructure to provide complementary supervision across heterogeneous tasks and jointly build transferable generation capabilities.
Our framework coordinates two complementary components, namely \textbf{capability-specific data pipeline} and \textbf{capability-aligned curriculum scheduling strategy}.
As shown in Figure~\ref{fig:data_framework}, we design specialized data engines for capability-oriented task separation, and stage-wise data stratification with an active refinement loop.
Together, these two components form a unified data infrastructure in which specialized data engines tailor supervision to individual capabilities and their mutual dependency, while the training curriculum dynamically composes their outputs as the model's capabilities evolve.

We first propose a capability-specific data pipeline comprising three interoperable engines for text-to-image generation, image editing, and knowledge-grounded generation.
Together, they instantiate complementary forms of relational visual supervision to develop multi-dimension capabilities.
The \textbf{T2I data engine} builds text-image grounding by expanding concept coverage, rebalancing long-tailed distributions, and aligning images with captions at multiple levels of granularity.
The \textbf{image-editing data engine} constructs supervision through a suite of specialized editing pipelines tailored to different editing operation types, while incorporating realistic associations mined from naturally related images and expert-generated examples for sparsely covered tasks.
The \textbf{knowledge-grounded data engine} links visual patterns to named entities and structured knowledge through knowledge-graph-guided acquisition.
This capability-oriented data organization enables independent measurement and optimization of specific capabilities, facilitating cross-task transfer.
Moreover, each data type covers a distinct subset of visual concepts, eliminating the need for every task-specific dataset to exhaustively cover all concepts.
For example, text-rendering capabilities acquired from synthetic T2I supervision can transfer to image editing, reducing the need to duplicate the same concept coverage in task-specific editing data~\citep{tuo2024anytext}.

Furthermore, specialization does not imply isolation.
A shared data-wrangling infrastructure and annotation conventions make the engine outputs interoperable, allowing visual concepts introduced in one pathway to be reused by another~\citep{UniVG,tian2025mige}.
In particular, we align editing instructions with T2I captions in both visual vocabulary and descriptive structure.
Moreover, we develop dense captions as precise supervision to accelerate training convergence on text-rich or structurally complex images.
Concepts already covered by T2I data can thereby transfer to editing supervision for better training convergence.

As for capability-aligned curriculum scheduling, we design a five-stage curriculum following the dependency order of capability acquisition.
The data \textbf{\emph{evolves along four coupled axes, including task composition, visual concept distribution, data quality, and image resolution}}, jointly aligned with the learning trajectory of our model.
Training begins with large-scale T2I data to establish broad visual-semantic alignment and basic generation, then incorporates structurally complex, knowledge-grounded, and text-rich examples to develop capabilities in structural composition, knowledge grounding, and text rendering.
Once a stable T2I prior has emerged, editing data adds reference preservation and controlled transformation while reusing the visual concepts already acquired from T2I supervision.
Continual training (CT) and supervised fine-tuning (SFT) subsequently shift the mixture toward smaller, balanced, and visually refined subsets, with image resolution scaled to content complexity so that additional computation is matched by richer supervision.
Notably, the schedule is not prescribed once and held fixed.
Capability-aware evaluation maps observed failures to targeted retrieval, expert construction, and resampling in the corresponding engines, and the refined data are incorporated into subsequent mixtures according to the model's current capability profile.

By coupling specialized supervision construction with dependency-aware curriculum scheduling, the framework promotes transfer across tasks and training stages, and turns otherwise isolated datasets into an adaptive data infrastructure for generalist image generation.

Our contributions are threefold.
\begin{itemize}[leftmargin=*]
  \item We introduce a \textbf{capability-specific data pipeline} that separates data construction by target capability while preserving transfer through shared data preprocessing and caption experts. Three specialized engines expand long-tail and defect-aware T2I coverage, mine natural visual associations for realistic editing supervision, and ground generation in structured knowledge. At scale, the pipeline curates a 440M-image T2I corpus from a billion-scale pool, over 120M high-quality image-editing pairs, and approximately 27M image-entity pairs for structured knowledge.
  \item We propose a \textbf{capability-aligned curriculum scheduling} strategy that follows the dependency order of capability acquisition rather than maintaining a fixed data mixture. Its multi-stage schedule jointly evolves task composition, visual-concept distribution, data quality, and resolution from 256px T2I pre-training to 1024px supervised fine-tuning, while capability-aware evaluation feeds residual gaps back into targeted retrieval, expert construction, and adaptive resampling.
  \item We develop a \textbf{captioning framework that bridges tasks and granularities}. Designed for billion-scale annotation, VLM-based captioner aligns editing instructions with the vocabulary and descriptive structure of T2I captions, provides multi-style supervision from entity tags to long-form descriptions, and generates verified dense captions for structured and text-rich images. This shared language interface promotes concept transfer across generation and editing while providing precise supervision from coarse semantics to fine-grained visual structure.
\end{itemize}

\section{Related Work}

\subsection{Large-Scale Visual Data Curation}

Web-scale image-text corpora, including LAION-5B, COYO-700M, and MMC4, have provided the data foundation for large-scale visual representation learning and image generation~\citep{laion,coyo,mmc4}.
Subsequent work has established data design as a central scaling dimension rather than a by-product of model training.
DataComp systematically studies filtering and selection over a fixed candidate pool~\citep{gadre2023datacomp}, while controlled scaling analyses show that data quality, semantic diversity, and text-conditioning density materially affect alignment and sample efficiency~\citep{li2024scalingt2i,lai2025revisitcaption,chen2026scalingtext}.
Another line of work improves supervision by replacing noisy web alt-text with synthetic descriptions.
DALL-E~3 demonstrates that highly descriptive captions substantially improve prompt following~\citep{dalle3}, and Recap-DataComp-1B scales VLM-based recaptioning to over one billion web images~\citep{li2024recap}.
Recent systems such as Qwen-Image and Seedream further integrate large-scale collection, filtering, recaptioning, semantic balancing, and targeted construction into end-to-end data pipelines~\citep{wu2025qwen,seedream2025seedream,zhao2026qwen}.
Knowledge-aware multimodal datasets additionally associate visual observations with named entities and structured facts for grounding and reasoning~\citep{gong2024uknow}.

These efforts substantially improve the quality and coverage of visual corpora.
Our work further shifts the unit of data design from the corpus to the capability.
We separate T2I, image-editing, and knowledge-grounded supervision into specialized yet interoperable engines, while shared semantic metadata and annotation interfaces allow concepts acquired in one pathway to support another.
This formulation makes capability-specific coverage gaps explicit without requiring each task-specific dataset to reproduce the full visual-concept distribution.

\subsection{Image-Editing Data Curation}

Instruction-based image editing is commonly learned from triplets comprising a source image, an editing instruction, and a target image.
Because naturally paired triplets are scarce, existing datasets largely scale supervision through synthesized transformations, model-generated targets, and automatic quality filtering~\citep{hui2024hq,chen2025sharegpt,pico-banana-400k}.
AnyEdit expands this paradigm with a fine-grained editing taxonomy and task-adaptive construction pipelines~\citep{yu2025anyedit}, while OmniEdit distills supervision from task specialists to cover diverse editing operations~\citep{wei2025omniedit}.
DreamOmni similarly constructs accurate editing pairs with operation-specific synthesis for unified generation and editing~\citep{xia2025dreamomni}.
These approaches have substantially improved the scale and diversity of editing data, yet their supervision remains dominated by transformations produced within a synthetic pipeline, which may simplify real-world relations or inherit artifacts from the generator.
Complementary studies derive manipulation cues from naturally occurring observations, for example by learning image transformations from temporal changes in videos~\citep{cao2025instructmove}.

In parallel, generalist models increasingly unify T2I generation and instruction-based editing within a shared architecture and training objective~\citep{xiao2025omnigen,xia2025dreamomni,UniVG,tian2025mige,seedream2025seedream}.
Efficient supervision across different tasks is required.
Thus, we combine operation-specific construction with editing relations mined from naturally associated images and expert-generated examples for sparsely covered tasks.
Meanwhile, editing instructions inherit the visual vocabulary and descriptive structure of T2I captions, allowing the editing corpus to focus on reference preservation and transformation rather than duplicating visual concepts already established by T2I supervision.

\subsection{Curriculum Learning and Adaptive Data Scheduling}

Curriculum learning organizes training examples in a meaningful order so that simpler concepts provide a foundation for learning more complex ones~\citep{bengio2009curriculum}.
Beyond example ordering, data-mixture optimization studies how training distributions should be composed across domains.
DoReMi, for instance, uses a proxy model and distributionally robust optimization to estimate domain weights for large-scale pre-training~\citep{xie2023doremi}.
Modern image-generation systems also employ multi-stage recipes that progressively vary resolution, data quality, task composition, or content complexity~\citep{wu2025qwen,team2025longcat,cai2025z,team2026firered,zhao2026qwen}.
Studies of unified multimodal pre-training further show that different generative and understanding capabilities emerge at different points in training rather than appearing simultaneously~\citep{bagel2025}.
Based on existing curricula design, our scheduling jointly evolves task composition, concept distribution, data quality, and resolution along the capability acquisition order, while capability-aware evaluation guides targeted data construction and resampling for subsequent stages.

\begin{figure}[t]
\centering
  \includegraphics[width=0.8\textwidth]{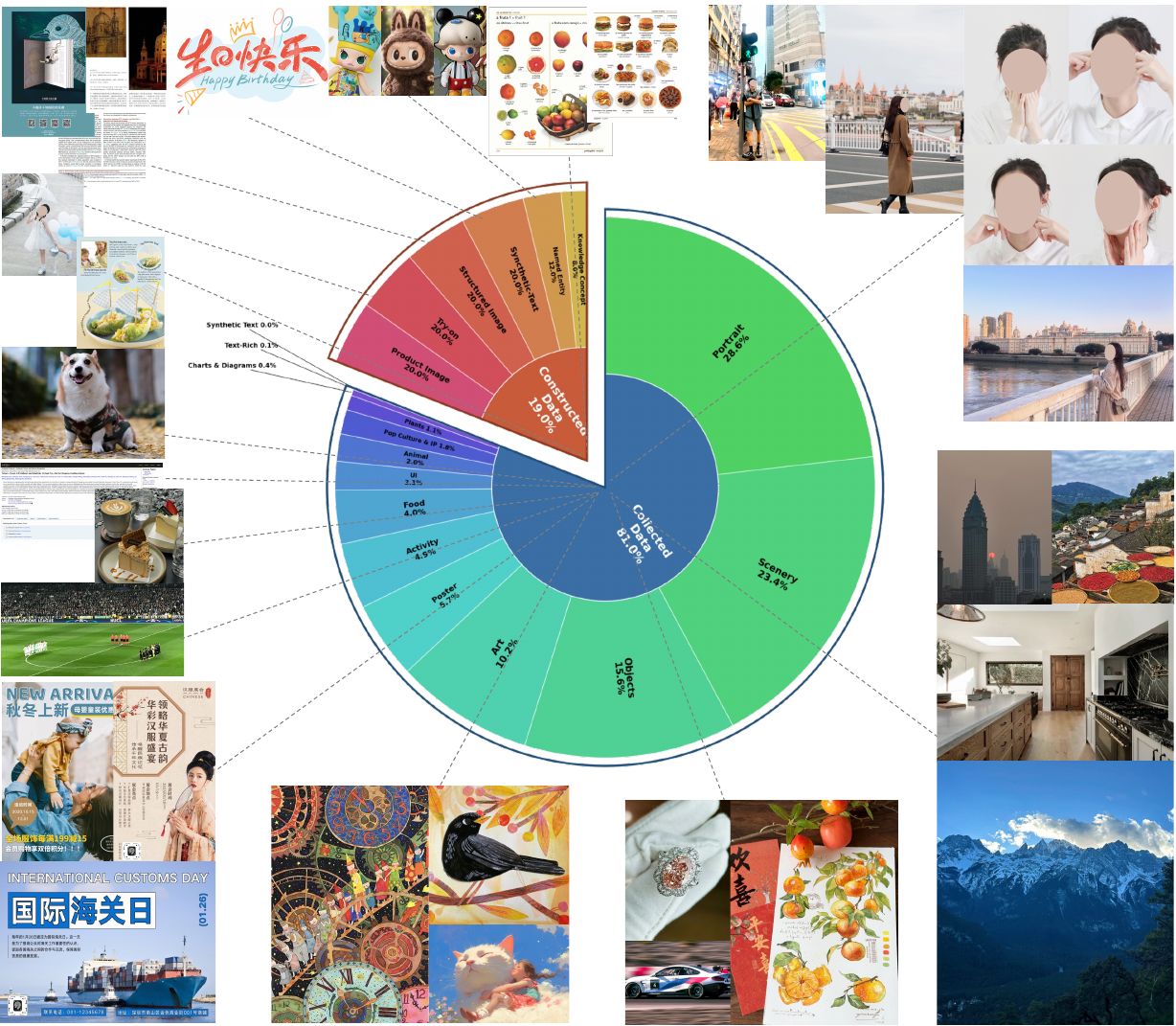}
  \caption{Distribution of the curated T2I corpus. The inner ring distinguishes collected and constructed data, while the outer ring reports the composition of visual domains retained for training.}
  \label{fig:t2i_data_dist}
\end{figure}

\section{Capability-specific Data Pipeline}
\label{sec:data_pipeline}

Motivated by the observation that capabilities learned from heterogeneous tasks can transfer across tasks~\citep{seedream2025seedream,UniVG}, we partition data construction into three specialized yet interoperable engines for T2I generation, image editing, and knowledge-grounded generation.
Each engine adopts construction and annotation mechanisms tailored to its target capability, while shared processing and supervision interfaces preserve concept transfer across tasks.
This organization makes capability-specific deficiencies independently measurable and actionable without requiring every task-specific dataset to exhaustively cover the same visual concepts.
Notably, the engines share the same data wrangling process, which illustrated in Appendix~\ref{sec:shared_wrangling}.

\subsection{T2I Data Engine for Visual-Semantic Grounding}

We develop a scalable T2I data engine to transform a billion-scale image pool into a high-quality and distribution-balanced training corpus of 440 million images.
The engine jointly optimizes data quality, semantic diversity, text-image alignment, and long-tail concept coverage.

\noindent\textbf{Scaling the T2I Corpus.}
We curate billions of noisy image-text pairs from heterogeneous sources, including public datasets~\citep{laion,coyo,mmc4}, image-rich websites, web-search engines, and e-commerce platforms.
All samples are processed by the shared data-wrangling pipeline described in Appendix~\ref{sec:shared_wrangling}.
Aesthetic quality, visual clarity, and AIGC detection serve as important filtering dimensions for T2I data~\citep{seedream2025seedream,zhao2026qwen}.
Each retained image is associated with semantic tags, quality attributes, and provenance information, allowing subsequent training stages to construct source-aware and distribution-aware mixtures.

\noindent\textbf{Capability-Oriented Coverage Expansion.}
T2I pre-training determines not only fundamental generation quality but also the visual concepts and world knowledge available to downstream capabilities~\citep{cai2025z,team2025longcat}.
We therefore expand T2I coverage along two complementary dimensions, namely entity-level concept coverage and coverage of both desirable and undesirable visual patterns.
For the former, we develop proactive acquisition pipelines that retrieve authentic user-created and professionally designed images, as well as targeted samples from long-tail visual domains, rather than relying solely on the natural distribution of web-scale corpora.
We additionally construct specialized collections of text-rich images, web pages, graphic designs, posters, social-media graphics, and commercial product images.
These collections establish visual-semantic correspondences for rare entities and structured visual content during T2I pre-training, which can subsequently be activated by downstream tasks.
For the latter, we avoid overly aggressive aesthetic filtering and retain a controlled proportion of imperfect images while explicitly describing their defects in captions.
This treatment turns otherwise discarded artifacts into identifiable visual concepts, enabling the model to recognize and avoid reproducing them.
Figure~\ref{fig:t2i_data_dist} summarizes the resulting T2I distribution.

\subsection{Image-Editing Data Engine for Relational Supervision}
 
\begin{figure}[t]
\centering
  \includegraphics[width=1.0\textwidth]{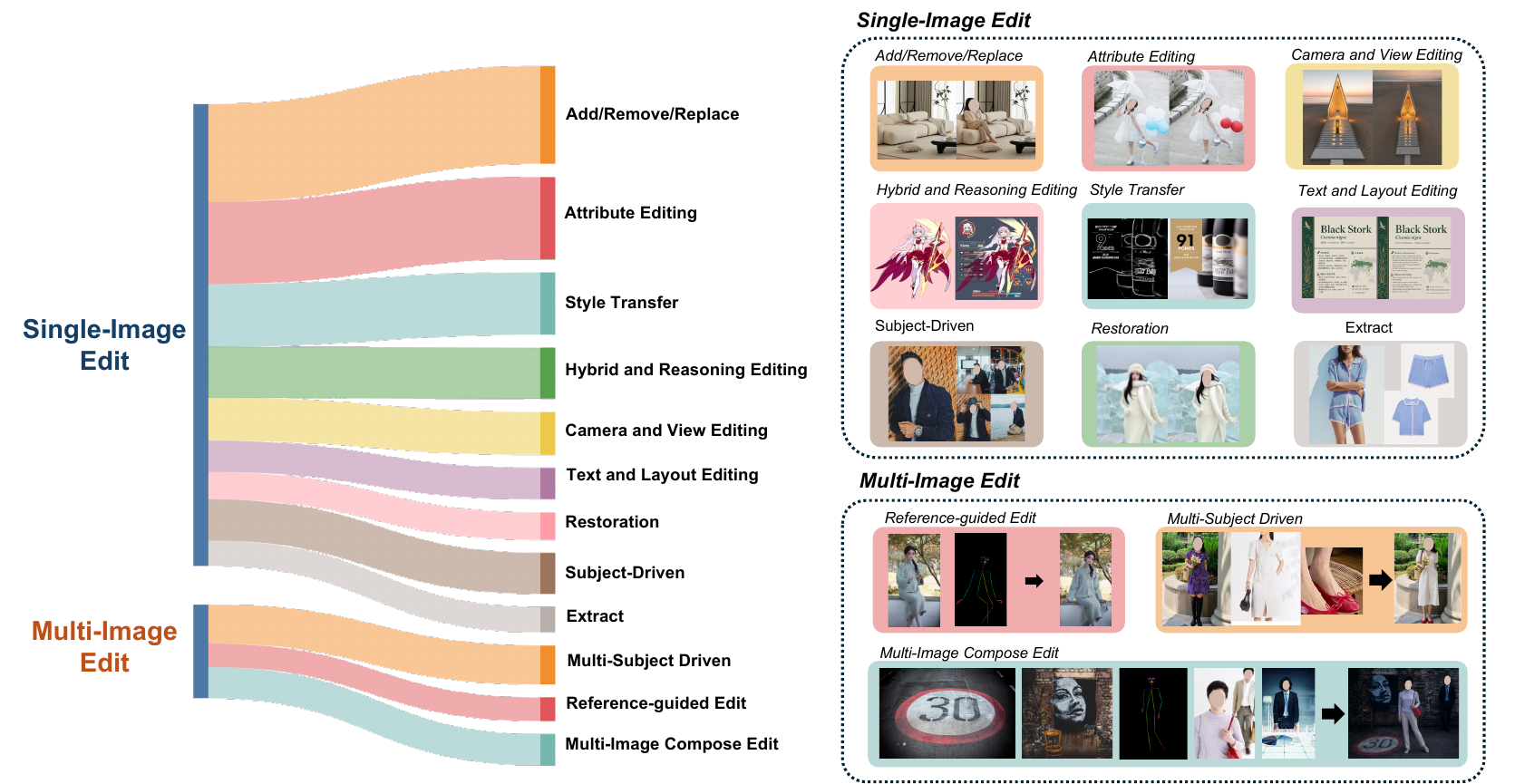}
  \caption{Composition of the image-editing corpus. The left Sankey diagram shows the relative mixture of single-image and multi-image editing tasks, and the right panels illustrate representative task families.}
  \label{fig:i2i_data_dist}
\end{figure}

High-quality image-editing pairs are substantially scarcer than unpaired images.
Existing studies primarily scale editing supervision through synthetically generated transformations~\citep{hui2024hq,pico-banana-400k,chen2025sharegpt}, which can oversimplify real-world changes and inherit artifacts from the generation pipeline.
We construct a mixture of complementary pipelines that recover editing associations from both actively constructed operations and naturally associated images.
All collected pairs are processed by the shared data-wrangling pipeline, rebalanced across operation types and image tags, and annotated by the editing-instruction captioner described in Sec.~\ref{sec:caption_interface}.
In total, the engine constructs 120M editing pairs spanning both single-image and multi-image editing tasks, as summarized in Figure~\ref{fig:i2i_data_dist}.

\noindent\textbf{Operation-Specific Pair Construction.}
We construct basic editing pairs by reversing observable changes in high-quality images.
A VLM first decomposes the scene and selects editable subjects, while SAM3~\citep{carion2025sam3} provides object-level masks for localized manipulation.
Starting from an image containing a target object, we remove or replace the selected region to obtain its counterpart.
Reversing the source and target order naturally produces complementary addition, removal, replacement, and attribute-modification pairs.
We also incorporate reference imitation~\citep{chen2024zero} to construct reference-conditioned supervision.

\noindent\textbf{Mining Natural Visual Associations.}
Purely synthesized pairs often make edited content appear pasted onto the input and fail to capture realistic combinations of transformations.
To improve realism, we mine editing relations from naturally associated images.
Explicit associations are predefined or actively retrieved, such as images of the same person, product, or object identified through visual embeddings and textual metadata.
Implicit associations are discovered from co-occurring images on the same web page, temporally adjacent video frames, e-commerce collections, and social-media posts.
Candidate pairs are retained only when identity consistency and semantic relevance are sufficiently high and a meaningful visual change is present.

We further recover latent relations from multi-panel layouts and composite images common in e-commerce and design data.
After splitting a composite image into individual regions, a VLM identifies the relation between panels.
This process recovers naturally paired examples such as product states before and after use, garments or cosmetics before and after application, product bundles and their components, and different states of the same object.
Compared with task-specific synthesis, these pairs provide realistic and hybrid editing operations that better reflect practical application scenarios.

\noindent\textbf{Expert Expansion and Quantitative Supervision.}
For difficult or underrepresented tasks, we first curate a small amount of domain-specific data and fine-tune base models such as Qwen-Image-2512~\citep{wu2025qwen} or FLUX.2~\citep{blackforest2025flux2klein} into task experts.
These experts generate candidate pairs for virtual try-on~\citep{chen2024wear}, beauty and makeup editing, product marketing image creation, reference-based editing, and other specialized tasks.
Video experts such as Wan~2.2~\citep{wan2025} are also used to generate controllable paired samples for challenging editing tasks~\citep{xu2024tunnel,chen2024livephoto,yao2025beyond,zheng2026itryon,song2026fashionchameleon,sun2026tryoncrafter}.
Candidate outputs are filtered according to instruction alignment, reference-identity preservation, and AIGC likelihood, converting limited high-quality supervision into a larger task-specific corpus.

We additionally construct pairs for quantitative transformations and visual-perception tasks.
The former include controlled object displacement, camera and photographic parameter adjustment, local color modification, and changes to text attributes.
For the latter, we form bidirectional pairs between RGB images and structural representations, including depth, edge, normal, and human-pose maps.
These data provide supervision for spatially and numerically precise editing.

\subsection{Knowledge-Grounded Data Engine for Entity and Structured Knowledge}
\label{sec:knowledge_engine}

To strengthen visual grounding and reasoning over knowledge-intensive entities, we curate knowledge-grounded visual data through two complementary pipelines, namely image-centric acquisition with fine-grained labeling and entity-centric retrieval guided by structured knowledge graphs.

\noindent\textbf{Knowledge-Graph-Guided Entity Acquisition.}
Starting from over 100 million Wikidata entities, we compute PageRank over the entity hyperlink graph to measure conceptual prominence and discard negligible-score candidates.
A VLM further assesses whether each remaining entity constitutes a meaningful knowledge concept and estimates its public recognition level.
This process yields approximately 3 million high-salience entity names.
For each entity, we retrieve web images and apply VLM-based visual-referential alignment to retain accurate depictions.
The same pipeline also captures structured knowledge representations, including theorem diagrams, mechanistic explanations, and scientific processes.
These data establish associations between visual patterns, real-world entities, and structured knowledge, supporting knowledge-intensive generation and complex editing scenarios that require abstract or structured reasoning.

\noindent\textbf{Image-Centric Entity Curation.}
Based on established entity set, we collect images from diverse online sources and apply domain-aware filtering to retain clearly identifiable entities across five major categories, including celebrities, landmarks, plants, animals, and popular IPs.
After stringent quality filtering, we obtain over 27 million high-quality image-entity pairs covering common knowledge concepts.

\subsection{Caption as Bridge to Align Across Tasks and Learn Across Granularities}
\label{sec:caption_interface}

We regard T2I captions and editing instructions as shared supervision interfaces that connect tasks and organize visual concepts across levels of granularity.
Across tasks, editing instructions inherit the visual vocabulary and descriptive structure of T2I captions, allowing concepts learned from T2I generation to transfer to reference-based editing.
Across granularities, the same image is annotated from entity-level concepts and concise prompts to long-form and dense descriptions, associating coarse intent with fine-grained visual control.

\noindent\textbf{Multi-Granularity T2I Recaptioning.}
The original metadata obtained during data collection is inadequate for learning fine-grained correspondences between language and visual elements.
Prior work has shown that descriptive synthetic captions substantially improve text-image alignment and prompt following~\citep{dalle3}.
We therefore recaption each curated image, first producing a comprehensive annotation that covers all visually grounded content.
Beyond the main subjects and their attributes, the annotation describes global composition, photographic and artistic style, illumination, color, fine-grained entities, spatial relations, relative subject scales, and visible OCR text when applicable.

\begin{figure}[t]
\centering
  \includegraphics[width=1.0\textwidth]{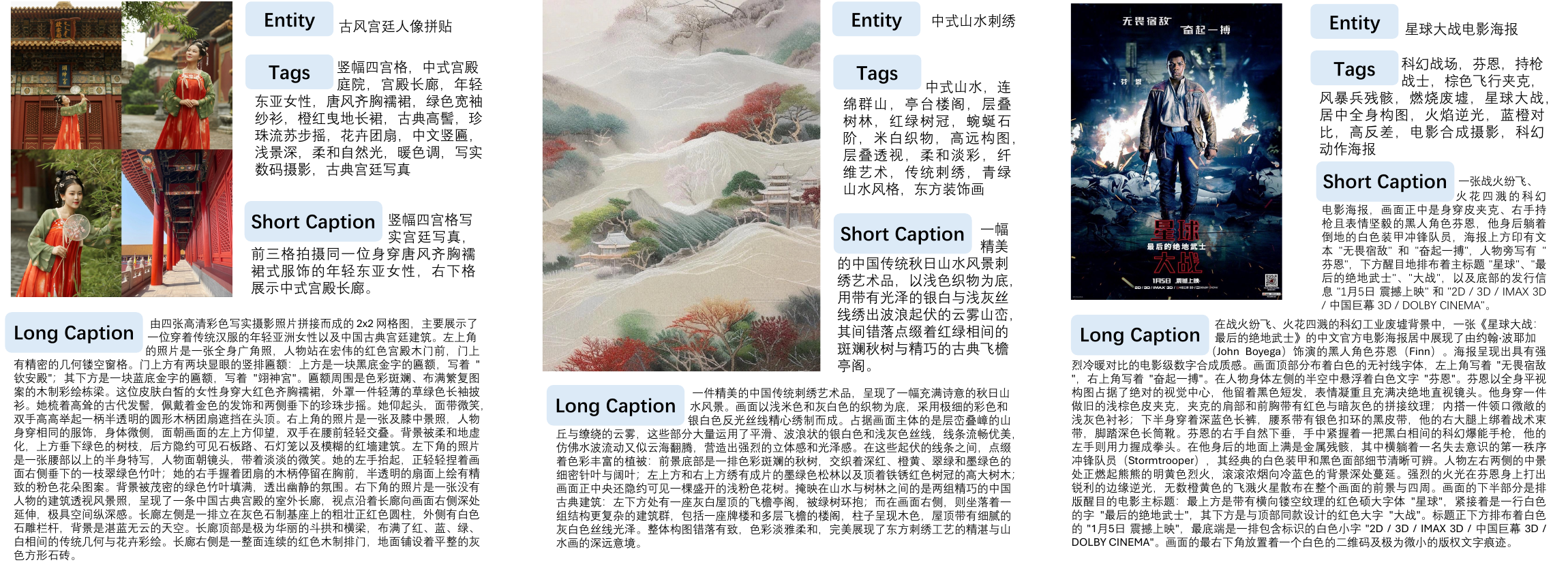}
  \caption{Multi-granularity T2I supervision. Starting from a comprehensive annotation of an image, we construct entity descriptions, tags, short prompts, and long-form captions at multiple levels of details.}
  \label{fig:data_t2i_caption}
\end{figure}

Starting from the comprehensive annotation and the image itself, we construct captions with multiple granularities and expressions, including entity descriptions, unordered tags, short prompts, medium-length captions, long captions, and dense descriptions, as shown in Figure~\ref{fig:data_t2i_caption}.
We additionally construct both Chinese and English variants.
Sampling across these caption styles exposes the model to heterogeneous prompting patterns and improves the diversity of text-image alignment.

\begin{figure}[t]
\centering
  \includegraphics[width=1.0\textwidth]{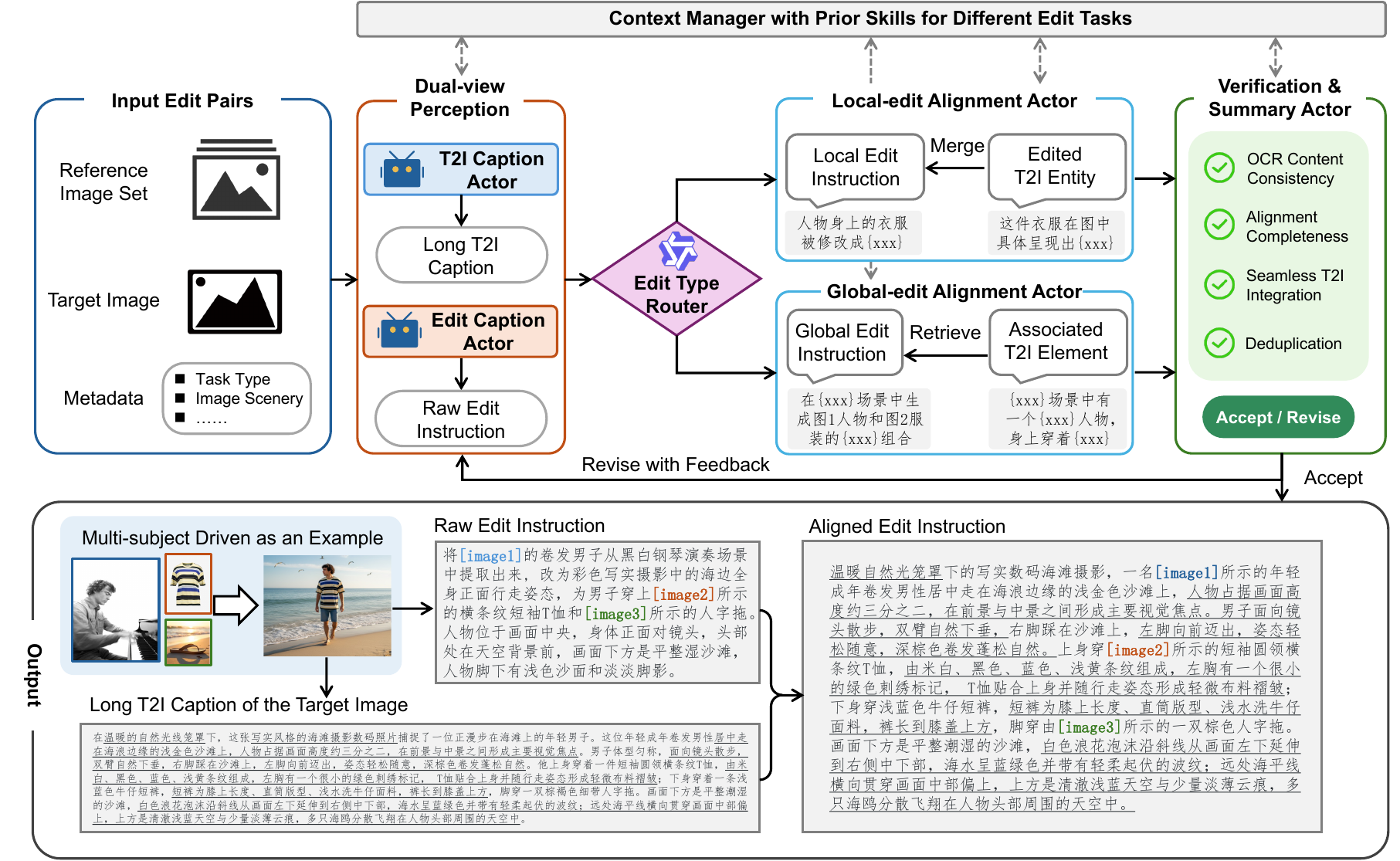}
  \caption{Pipeline for constructing T2I-aligned editing instructions. Dual-view perception produces a target-image caption and a raw editing instruction; task-specific actors align local or global changes with reusable T2I descriptions, followed by consistency verification and iterative revision.}
  \label{fig:data_edit_caption}
\end{figure}

\noindent\textbf{T2I-Aligned Editing Instructions.}
We train a VLM-based captioner to convert raw editing pairs into a unified supervision format.
As illustrated in Figure~\ref{fig:data_edit_caption}, we treat image editing as conditional T2I generation in which the target description inherits the vocabulary and descriptive style learned from T2I data, while the instruction additionally expresses image references, transformations, and preservation constraints.
The captioner first generates a reconstruction-level dense description of the target image without observing any source image.
It then jointly examines the source and target images through fine-grained comparison to identify their visual differences.
Regions or entities in the target description that can be directly preserved from a source image are replaced with the corresponding \texttt{[image N]} reference.
Visual information already present in a source is therefore expressed by reference rather than repeated in text, while only attributes that differ between the source and target remain as explicit descriptions.
Based on this aligned representation, we construct multiple instruction variants, including detailed editing instructions, concise commands, and simulated user requests, without changing the underlying visual transformation.

\noindent\textbf{Specialized Caption Experts.}
To perform billion-scale annotation with consistent quality, we develop two VLM-based caption experts, namely a general captioner and a dense captioner, initialized from Qwen3.5-27B.
We first apply prompt engineering to improve caption accuracy and granularity, incorporating comprehensive annotation-dimension design, self-reflection, and in-context learning.
We then collect high-quality T2I captions and editing instructions generated by strong teacher models to establish detailed and fluent captioning behavior.
The captioner is fine-tuned on a unified T2I and editing dataset covering diverse image and editing types; higher weights are assigned to difficult tokens, including spatial terms and OCR content.
We further apply reinforcement learning with rewards for visual-content coverage, hallucination, and linguistic clarity.
For text-rich images such as posters and product advertisements, long-form OCR accuracy is assessed jointly by VLM-based scoring and rule-based verification.
An anti-hacking reward penalizes subjective judgments, unsupported interpretations, redundant statements, and other content that does not contribute to training supervision.

Dense captions specify layouts, OCR text, and visible elements, allowing DiT models to learn associations between textual conditions and structured visual content~\citep{BizGen}.
For each image, the dense captioner first identifies its visual type and constructs a corresponding description outline.
Photographic images are organized from the main subjects to scene, composition, lighting, color, and background, whereas structured or text-rich images are organized by layout regions, text blocks, and decorative elements.
The captioner verifies object identities and counts, attributes, spatial relations, relative scales, OCR text, and peripheral content item by item, and expresses uncertain details at a safer level of specificity.
The verified elements are then assembled from global structure to local details.
To jointly improve accuracy and exhaustiveness, we train the dense captioner with a self-verification reasoning process followed by dense-caption generation.
The training traces exploit natural disagreements among multiple teacher models to construct trajectories of initial assessment, uncertainty, and correction, thereby internalizing reflection without relying on an external verifier at inference time.
After supervised fine-tuning, reinforcement learning with multi-dimensional rewards further optimizes factual accuracy, visual coverage, and structural organization.

\section{Capability-aligned Curriculum Scheduling}
\label{sec:curriculum}

We construct a multi-stage data curriculum spanning foundational pre-training, continual training, and supervised fine-tuning.
Rather than maintaining a fixed data mixture, the curriculum follows the dependency order of capability acquisition and evolves along four coupled axes, i.e., task composition, visual-concept distribution, data quality, and image resolution.
The pipeline first establishes broad semantic coverage from large-scale T2I data, then introduces structurally complex, knowledge-grounded, text-rich, and image-editing supervision, and finally transitions toward balanced and refined subsets.

\begin{figure}[t]
\centering
  \includegraphics[width=1.0\textwidth]{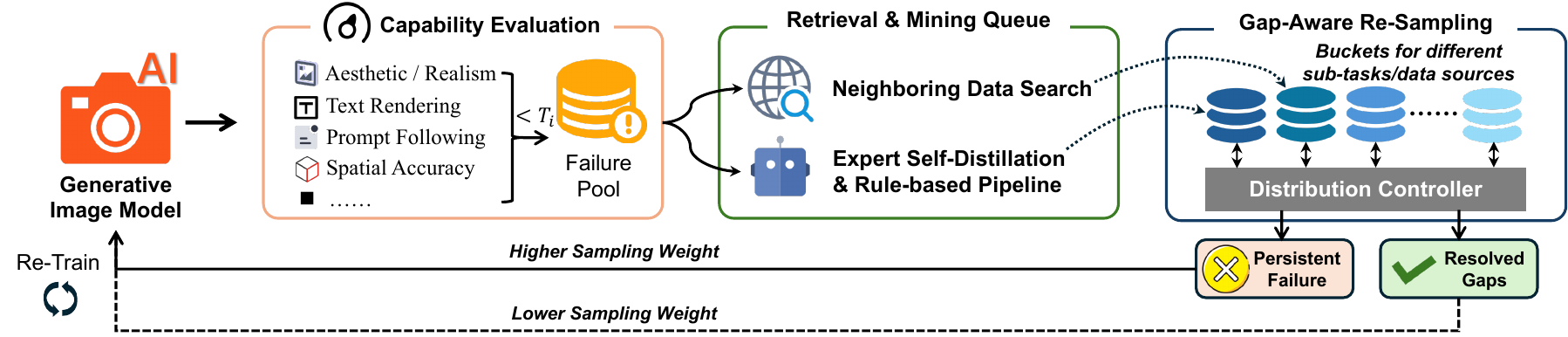}
  \caption{Capability-gap-driven active feedback loop. Capability-aware evaluation identifies failure cases, which seed neighboring-data retrieval and expert-driven construction. Gap-aware resampling then increases the weights of persistent failures and down-weights resolved gaps in subsequent training.}
  \label{fig:loop_framework}
\end{figure}

\subsection{Dependency-Aligned Multi-Stage Curriculum Strategy}

All samples produced by the capability-specific data engines are maintained in a shared data reservoir and indexed using multi-dimension attributes.
We use these attributes to determine the data clusters eligible at different stage and the corresponding sampling weight.

\paragraph{Stage 1: 256px T2I Pre-training.}
The first stage uses large-scale 256px T2I data to maximize semantic coverage and preserve the authentic long-tail distribution of real-world visual content.
We adopt an inclusive filtering strategy based primarily on image metadata and heuristic rules, avoiding aggressive aesthetic filtering that may remove visually imperfect but semantically useful samples such as old photographs.
With accurate captions, a controlled subset of images containing visual imperfections is retained so that the model can explicitly learn the distribution of such defects and avoid them in subsequent generation.
The corpus is organized into aspect-ratio buckets to prevent excessive cropping or deformation, and rare concepts receive moderately increased sampling weights.

\paragraph{Stage 2: 256px/512px Complex T2I Pre-training.}
Building on the broad T2I corpus from Stage~1, Stage~2 progressively extends the target resolution from 256px to 512px.
We introduce content whose visual structure cannot be modeled effectively at 256px, particularly dense text-rendering images, layout-sensitive samples, and knowledge-grounded visual content.
This stage couples increased resolution with increased content complexity to improve structural and detail fidelity.

\paragraph{Stage 3: Joint 512px T2I\&Edit Pre-training.}
After 512px T2I generation has stabilized, Stage~3 introduces both natural and synthetic editing pairs.
Editing samples are balanced across instruction categories, while the T2I branch preserves broad semantic and stylistic coverage.
The two sources are combined under a unified pre-training setting, allowing the model to reuse concepts acquired from T2I supervision while learning reference preservation and controlled transformation.

\paragraph{Stage 4: 512px/1024px T2I\&Edit Continual Training.}
During continual training (CT), the target resolution progressively increases from 512px to 1024px.
To improve visual quality while preserving world knowledge, the data distribution shifts from broad but noisy pre-training data toward cleaner and more visually refined sources.
We remove web-crawled sources with low quality bounds and increase sampling from high-fidelity sources and professional visual domains.
A VLM assigns multi-level semantic categories to control the distribution shift through global resampling and proportional balancing.
Editing data undergoes a second round of source filtering and is rebalanced across instruction categories to maintain stable coverage under multi-task training.

\paragraph{Stage 5: 1024px T2I\&Edit Supervised Fine-Tuning.}
The supervised fine-tuning (SFT) stage constructs a small-scale and highly curated dataset that guides the model toward a high-quality sub-manifold of the CT distribution with stronger visual fidelity and instruction alignment.
We sample from rigorously defined top-tier sources and apply a two-step review pipeline consisting of VLM-based preliminary screening followed by human re-evaluation.
Samples with visible defects or weak text-image alignment are removed, and global category balancing prevents overfitting to dominant domains and mitigates forgetting of long-tail concepts.

\subsection{Capability-Gap-Driven Active Feedback Loop}

A fixed data distribution cannot continuously match the evolving capabilities of the model.
We therefore treat data curation as an evaluation-driven optimization process in which evaluation results from preceding training stages identify underperforming task types and semantic concepts.
As shown in Figure~\ref{fig:loop_framework}, these capability gaps drive targeted data retrieval, expert construction, and gap-aware resampling for subsequent stages.

\begin{figure}[p]
    \centering
    \includegraphics[width=1.0\linewidth]{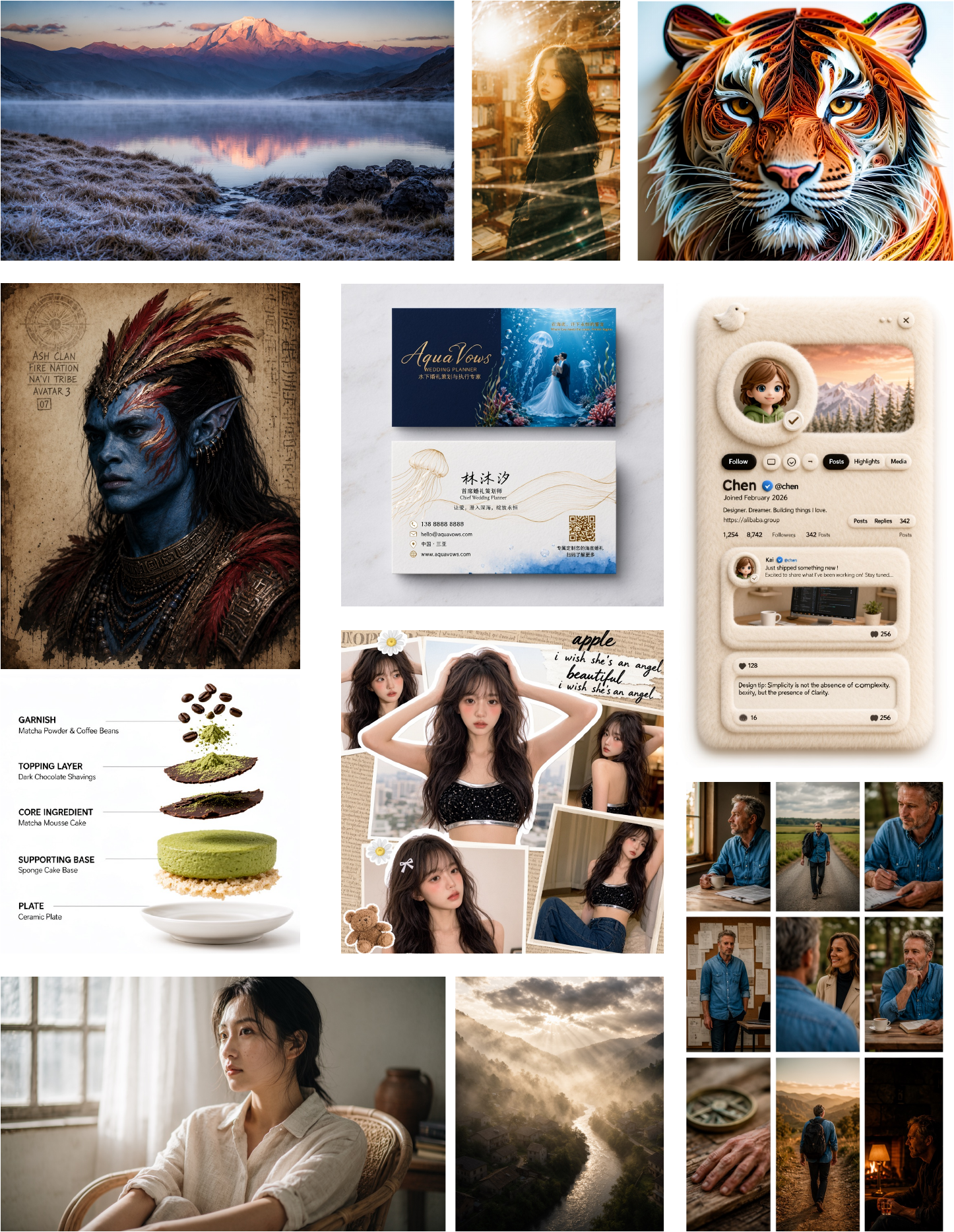}
    \caption{Qualitative T2I results across illustration, graphic design, knowledge visualization, portraiture, landscapes, multi panel composition, and photographic style control.}
    \label{fig:t2i_qualitative_results}
\end{figure}

\noindent\textbf{Capability-Aware Failure Discovery}
We assess each intermediate checkpoint with capability-stratified evaluation covering both T2I generation and image editing.
Each failed sample below its capability-wise quality threshold $T_i$ is annotated with its task type, hierarchical semantic tags, and failure dimensions.
This process converts individual failure cases into measurable capability gaps and prioritizes recurring failure modes.

\noindent\textbf{Distribution Update and Loop Closure}
Recurring failure modes attributable to data are used as retrieval seeds to search for or construct neighboring samples with diverse prompt formulations.
For T2I data, when suitable real data are insufficient, we invoke capability-specific construction pipelines or expert generation to synthesize candidates.
For editing data, especially underrepresented instruction types, we retrieve appropriate reference images and either mine naturally associated pairs or construct expert-generated pairs.
A subset of the newly added data undergoes VLM-based preliminary assessment followed by human review before entering the SFT pool.

Accepted supplementary samples are organized into task- and source-specific buckets, whose sampling weights are adjusted according to the capability gaps of the evolving model.
Buckets associated with persistent failures receive higher weights, whereas resolved gaps are down-weighted; unresolved cases are returned to the data-mining queue.
In this way, evaluation is converted from a terminal measurement into an active control signal for continuously improving the T2I and editing distributions.

\begin{figure}[t]
    \centering
    \includegraphics[width=0.90\linewidth]{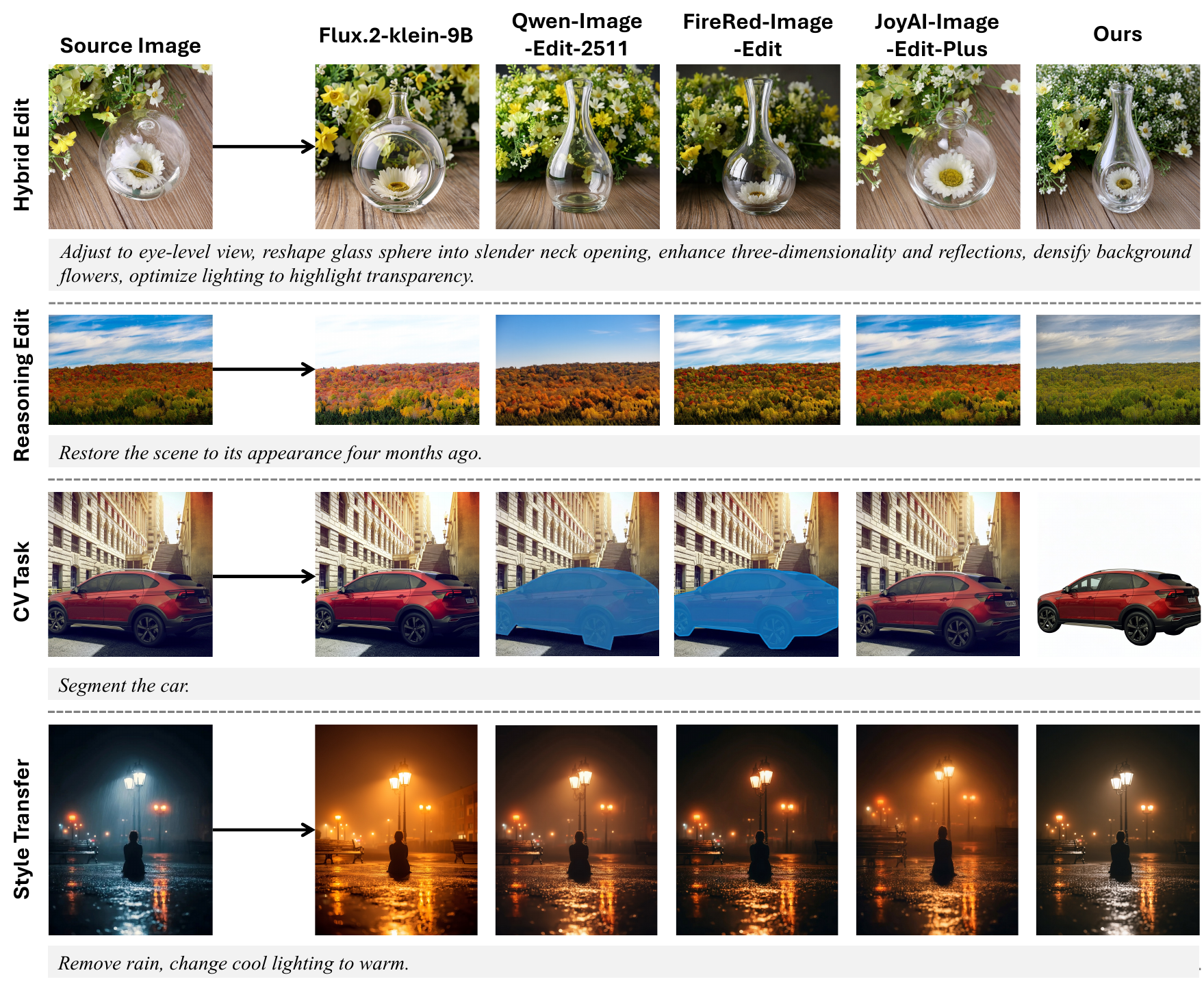}
    \caption{Qualitative comparison on challenging single image editing cases. The examples cover hybrid editing, reasoning based transformation, object segmentation, and style transfer.}
    \label{fig:single_image_edit_results}
\end{figure}

\begin{figure}[t]
    \centering
    \includegraphics[width=1.0\linewidth]{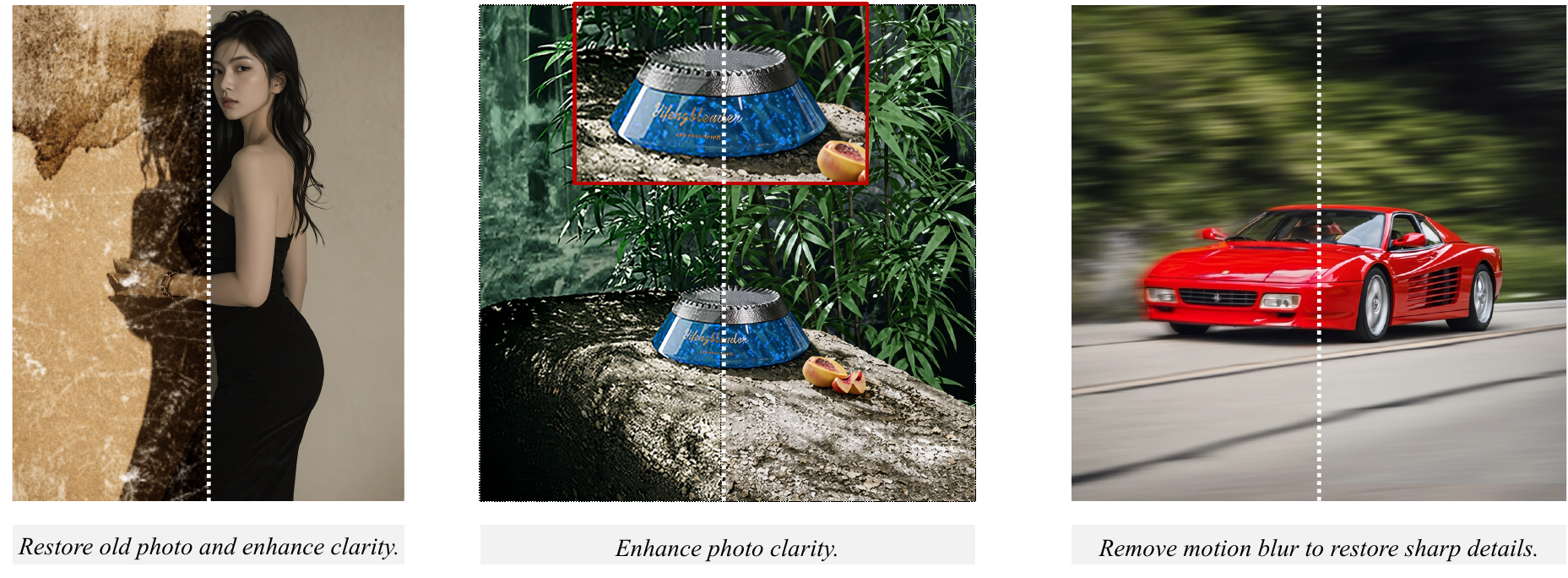}
    \caption{Visualization of degradation-aware restoration for different degradation types.}
    \label{fig:negative_aware_generation}
\end{figure}

\section{Experiments}
\label{sec:experiments}

\subsection{Text-to-Image Generation}
We present qualitative T2I results across a broad range of visual styles and formats in Figure~\ref{fig:t2i_qualitative_results}.
These results indicate that the model performs well in complex T2I generation, such as designed illustrations, multi-panel images, knowledge-structure visualization, and photographic style control.
Our scalable T2I data engine contributes this by producing images rich in text and structured layouts, which providing supervision for posters, interfaces, diagrams, and multi panel composition.
The capability aligned curriculum then introduces complex structure and higher resolution supervision after broad visual grounding, allowing diversity, compositional accuracy, and rendering quality to improve together.

\subsection{Image Editing.}
\noindent{\textbf{Quantitative Evaluation.}} We evaluate image editing on CPI-General-Bench and CPI-Practical-Bench, two subsets of CPI-Bench~\citep{zhou2026cpibench}.
CPI-Bench is a comprehensive, practical, and intelligent benchmark for image editing in real-world settings and comprises three complementary subsets.
Specifically, CPI-General-Bench provides broad coverage of fundamental editing capabilities, including CPI-Practical-Bench focuses on frequently encountered real-world application scenarios, and CPI-Intelligent-Bench evaluates editing instructions that require advanced reasoning.
CPI-General-Bench contains 2,039 examples spanning 30 fundamental tasks (20 single image and 10 multi image tasks).
CPI-Practical-Bench contains 558 examples covering 51 common application types across the four domains of portrait enhancement, electronic commerce and advertising creativity, residential and interior design, and content creation.
Using the proposed data pipeline and curriculum, we train MM-DiT models with 3B and 6B sizes, and evaluate on multiple tasks.
The evaluation results on CPI-General-Bench and CPI-Practical-Bench are shown in Table \ref{tab:editing_capability}.
Metrics are assessed via VLM across multiple distinct dimensions, as illustrated in \citep{zhou2026cpibench}, with scores ranging from 1 to 5.

\begin{table}[t]
    \centering
    \caption{Image editing quantitative evaluation on CPI-General-Bench and CPI-Practical-Bench. Overall denotes their arithmetic mean.\label{tab:editing_capability}}
    
    \setlength{\tabcolsep}{20pt}
    \renewcommand{\arraystretch}{1.5}
    \begin{tabular}{lcccc}
        \toprule
        \textbf{Model} & \textbf{Parameters} & \textbf{CPI-General} & \textbf{CPI-Practical} & \textbf{Overall$\uparrow$} \\
        \midrule
        Our Model-3B & 3B & 3.95 & 3.91 & 3.93 \\
        Our Model-6B & 6B & 3.96 & 3.92 & 3.94 \\
        \bottomrule
    \end{tabular}
\end{table}

\noindent{\textbf{Qualitative Evaluation.}}
We present challenging single-image editing cases in Figure~\ref{fig:single_image_edit_results}  that span hybrid transformation, reasoning editing, etc.
These cases require the model to infer the intended visual state and preserve content beyond direct appearance matching.
Such capabilities benefit from abundant editing pairs mined from natural sources, where the transformation reflects relationships that occur in everyday settings.
These naturally occurring relations provide realistic supervision to learn complex image transformations.

\begin{figure}[t]
    \centering
    \includegraphics[width=1.0\linewidth]{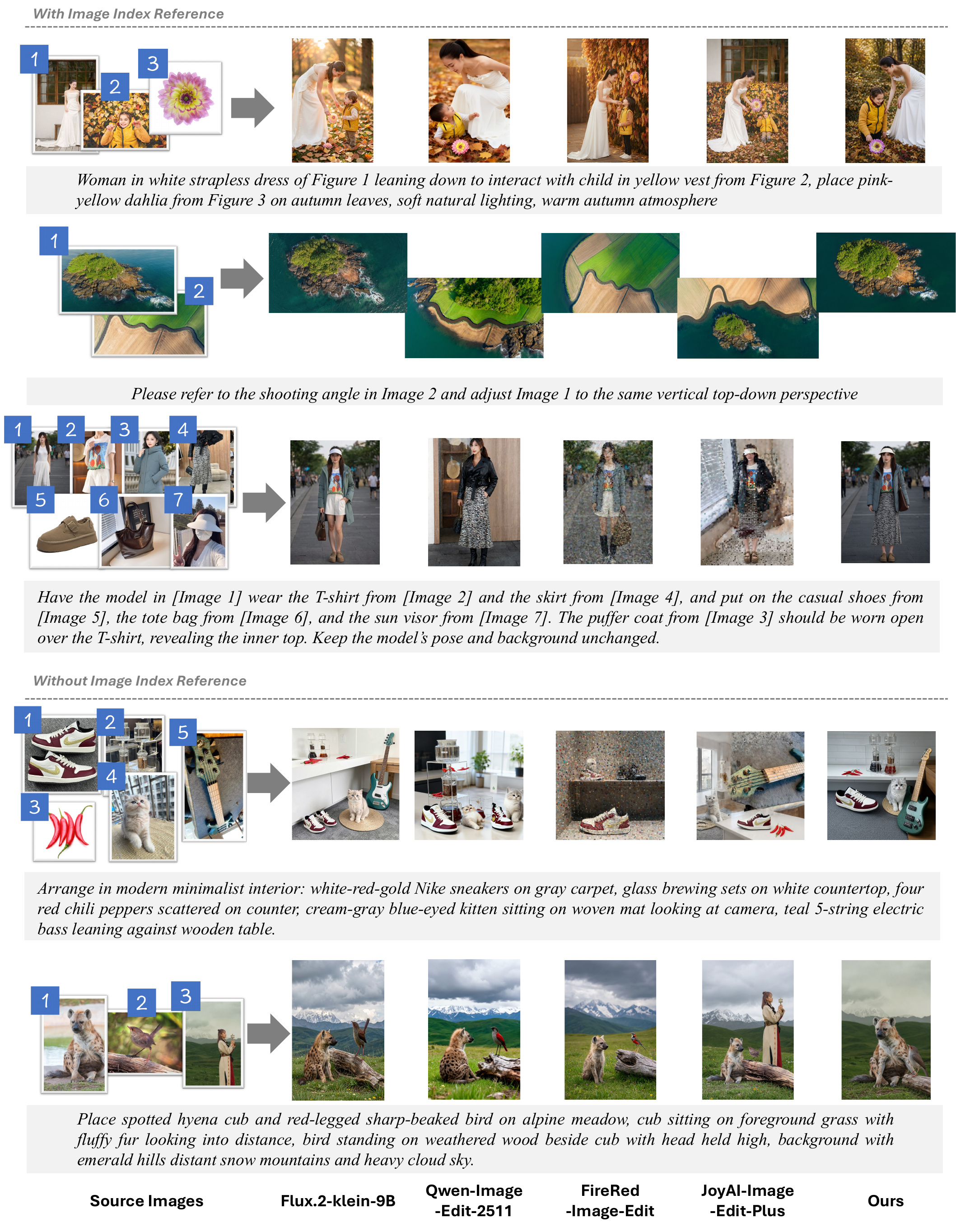}
    \caption{Qualitative comparison on multi image editing. The examples evaluate viewpoint alignment, reference relation understanding, and composition capabilities. Comparison cases are grouped by with/without explicit image index reference.}
    \label{fig:multi_image_edit_results}
\end{figure}

Figure~\ref{fig:negative_aware_generation} shows restoration from old photographs, low clarity, and motion blur.
Learning this behavior requires retaining a small and controlled portion of degraded images together with explicit descriptions of their degradation states.
This supervision enables the model to recognize defects in an input image, and also helps the model distinguish undesirable visual degradation from valid content, which supports higher generation quality.

\clearpage
Furthermore, we present complex cases involving viewpoint transfer, reference relation understanding, text rendering, and composition from multiple source images, as shown in Figure~\ref{fig:multi_image_edit_results}.
The model identifies which visual attributes belong to each reference and binds them to the corresponding parts of the instruction.
This requires image captions and editing instructions to maintain explicit correspondences between textual semantics and visual elements.
Explicit cross-task alignment allows the model to combine subjects, attributes, layouts, and rendered text without confusing their sources.

\section{Conclusion}
\label{sec:conclusion}

In this work, we present a capability-driven data infrastructure for generalist image generation and editing.
Three specialized yet interoperable engines construct complementary supervision for visual-text grounding, inter-image transformation, and image-knowledge association, while a shared captioning interface enables concept transfer across tasks and granularities.
A capability-aligned curriculum then jointly evolves task composition, visual-concept distribution, data quality, and image resolution, while evaluation-driven retrieval, expert construction, and resampling close the refinement loop.
At scale, the infrastructure curates a 440M-image T2I corpus, 120M editing pairs, and over 27M image-entity pairs, enabling the training of a MM-DiT model from scratch.
Qualitative results illustrate the breadth of generation and editing capabilities supported by the resulting models.
Overall, our study establishes data organization as a complementary scaling axis and reframes data curation as an adaptive supervision system rather than a collection of isolated task-specific pipelines.

\bibliography{references}

\clearpage

\appendix

\section*{Appendix}

\noindent\rule{\textwidth}{0.4pt}

\section{Shared Data Wrangling}
\label{sec:shared_wrangling}

Basic data-wrangling strategies are shared across the capability-specific data engines.
The shared pipeline standardizes data validity and quality, extracts comparable semantic metadata, and provides the attributes required for stage-specific filtering and rebalancing in the curriculum scheduling.

\subsection{Filtering and Quality Control}

\paragraph{Basic Rule Filtering.}
We first convert all images to RGB format and profile each sample using basic metadata.
Images that cannot be decoded because of file corruption are removed.
We impose a strict resolution lower bound and discard images containing fewer than $256^2$ total pixels.
Samples with extreme aspect ratios are filtered according to stage-specific requirements.

\paragraph{Technical Quality Filtering.}
We develop dedicated pipelines to detect blur and severe compression.
Severe blur is detected by combining Laplacian variance with BRISQUE~\citep{mittal2012no} scores, while file entropy and JPEG-quality estimation reject images with excessive compression and pronounced block artifacts.
Pixel variance identifies solid-color or nearly blank images.
For white-background images common in e-commerce and stock-media data, we combine RGB entropy with the proportion of black and white pixels for controlled downsampling, preserving their conceptual value while preventing them from dominating the pre-training distribution.

\paragraph{Perceptual Quality Filtering.}
We train clarity and aesthetic predictors to evaluate visual clarity and perceptual quality.
The filtering thresholds vary across training stages so that early pre-training preserves semantic diversity while later stages progressively emphasize visual quality.

\paragraph{Deduplication.}
We employ a three-level deduplication pipeline spanning exact, near-duplicate, and semantic matching.
MD5 and file hashes first remove identical samples at low computational cost.
pHash then detects near-duplicate images produced by minor cropping, watermarking, or resizing.
Finally, DINOv3~\citep{dinov3_2025} extracts image-level embeddings for high-dimensional clustering with FAISS~\citep{douze2025faiss}.
For samples whose intra-cluster cosine similarity exceeds $0.99$, we retain the highest-quality image as determined jointly by resolution and sharpness.

\paragraph{Watermark and Text Detection.}
Specialized detectors identify watermarks, logos, subtitles, and overlaid text.
Images dominated by watermarks and text-rich images are marked separately for subsequent filtering and captioning rather than treated as a single category.

\paragraph{AIGC Detection and Content Safety.}
AI-generated images may contain latent artifacts that limit the upper bound of generation quality.
We train an AIGC classifier to identify synthetic images in naturally sourced corpora and remove high-confidence AIGC samples from the pre-training pool.
An NSFW detector and metadata-based unsafe-keyword filtering are additionally applied for content safety.

\subsection{Hierarchical Metadata Extraction and Rebalancing}

After filtering, the remaining web-scale pool still exhibits a long-tailed semantic distribution dominated by frequent concepts.
We therefore extract hierarchical semantic metadata and rebalance the data with schedules tailored to different training stages.

\paragraph{Taxonomy Construction.}
We adopt the leaf nodes of an established visual classification hierarchy as over 280K fine-grained semantic tags.
These tags are organized into a four-level taxonomy whose three upper levels contain 15, 74, and 331 categories, respectively.
Each fine-grained tag is mapped to a leaf node, enabling both coarse- and fine-grained distribution control.

\paragraph{Tag Assignment.}
For each image, we compare its caption embedding with tag embeddings by cosine similarity to retrieve the top-1000 candidate tags.
An adaptive filter combines semantic similarity with hierarchical relations to retain up to 15 representative and semantically diverse tags per image, providing compact metadata across multiple conceptual dimensions.

\paragraph{Data Rebalancing.}
We perform hierarchical resampling according to two principles.
First, all semantic tags are represented in the final corpus, with additional attention to rare and long-tail concepts.
Second, sample counts are approximately balanced across first-level categories and recursively among child categories under the same parent down to the third level.
This hierarchical strategy reduces the dominance of frequent concepts while preserving semantic diversity at multiple granularities.

\end{document}